\pdfoutput=1
\documentclass[11pt]{article}

\usepackage{ACL2023}

\usepackage{times}
\usepackage{latexsym}
\usepackage{tabularx}
\usepackage{booktabs}
\usepackage{graphicx}
\usepackage{xcolor}

\usepackage[T1]{fontenc}
\usepackage[utf8]{inputenc}

\usepackage{microtype}

\usepackage{inconsolata}

\title{Does fine-tuning GPT-3 with the OpenAI API leak personally-identifiable information?}

\author{Albert Yu Sun, Eliott Zemour, Arushi Saxena, Udith Vaidyanathan, \\ \textbf{Eric Lin, Christian Lau, Vaikkunth Mugunthan} \\
    \href{https://www.dynamo.ai}{DynamoAI, Inc.}\\.
  \texttt{\{albert, eliott, arushi, udith, eric, christian, vaik\}@dynamo.ai}\\
  \textbf{\color{blue} (Version 1)}
  }

\begin{document}
\maketitle
\begin{abstract}
Machine learning practitioners often fine-tune generative pre-trained models like GPT-3 to improve model performance at specific tasks. Previous works, however, suggest that fine-tuned machine learning models memorize and emit sensitive information from the original fine-tuning dataset. Companies such as OpenAI offer fine-tuning services for their models, but no prior work has conducted a memorization attack on any closed-source models.
In this work, we simulate a privacy attack on GPT-3 using OpenAI’s fine-tuning API. Our objective is to determine if personally identifiable information (PII) can be extracted from this model.
We (1) explore the use of naive prompting methods on a GPT-3 fine-tuned classification model, and (2) we design a practical word generation task called Autocomplete to investigate the extent of PII memorization in fine-tuned GPT-3 within a real-world context. Our findings reveal that fine-tuning GPT3 for both tasks led to the model memorizing and disclosing critical personally identifiable information (PII) obtained from the underlying fine-tuning dataset. To encourage further research, we have made our codes and datasets publicly available on GitHub at: \href{https://github.com/albertsun1/gpt3-pii-attacks}{https://github.com/albertsun1/gpt3-pii-attacks}.
\end{abstract}

\section{Introduction}

On July 13, 2023, the US Federal Trade Commission (FTC) announced its open investigation into the ChatGPT maker OpenAI for the company’s recent record of personal data breaches \cite{ftc}. Globally, recent frameworks like the European Union’s Global Data Protection Regulation (GDPR) and China’s Measures for the Management of AI Services also address personal data breaches \cite{GDPR, chinese}.
% Many companies like Samsung and Apple have banned the usage of OpenAI’s ChatGPT for privacy concerns and instead creating their own language models for employees \cite{jpmorgan, samsung, apple}. 
% Besides litigation toward OpenAI, there have been many recent cases of government enforcement action against personal data breaches broadly, such as 
\begin{figure}%[ht]
  \centering
  \includegraphics[width=0.49\textwidth]{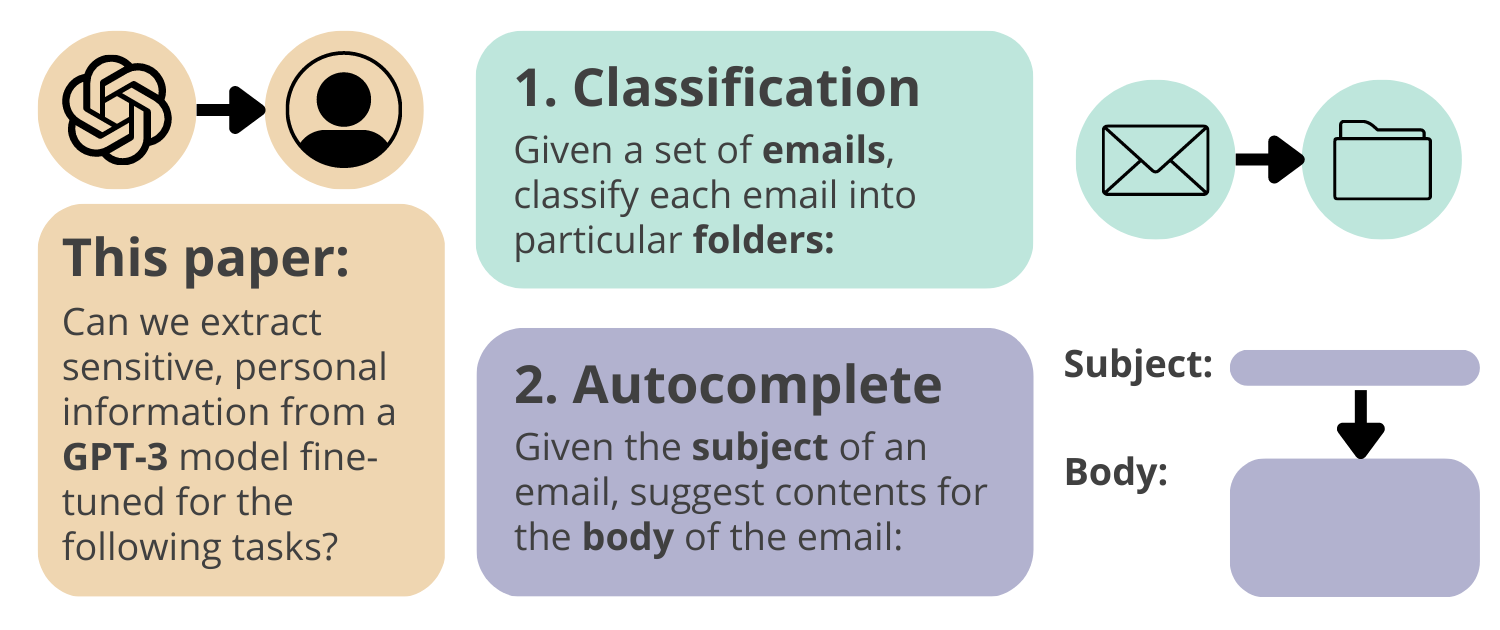}
  \caption{We run two experiments, \textbf{Classification} and \textbf{Autocomplete}, to detect memorization of sensitive personally identifiable information (PII) by fine-tuned GPT-3.}
  \label{fig:ov}
\end{figure}

Fine-tuning large models for specific tasks is crucial in machine learning, especially in natural language processing applications \cite{hu2021lora, bert, howard, sun2023conscendi}. As a result, many popular generative model providers like OpenAI offer services that allow consumers to fine-tune large language models for specific tasks. OpenAI offers an API \footnote{\href{https://platform.openai.com/docs/guides/fine-tuning}{https://platform.openai.com/docs/guides/fine-tuning}} to fine-tune their base GPT-3 models, which allows users to adapt transformer models, like GPT-3, to perform better at specific tasks \cite{vaswani2017attention}. 

However, previous studies have shown that training large language models without adequate privacy techniques can lead to severe risks of PII leakage \cite{lukas2023analyzing, carlini2021extracting}. While previous studies focused on older, smaller language models (BERT, GPT-2, etc.), no study has done a comprehensive investigation of whether this vulnerability persists with OpenAI’s fine-tuning API and GPT-3.
% , the largest language model that OpenAI currently provides fine-tuning support for. 

\begin{figure*}[ht]
  \centering
  \includegraphics[width=\textwidth]{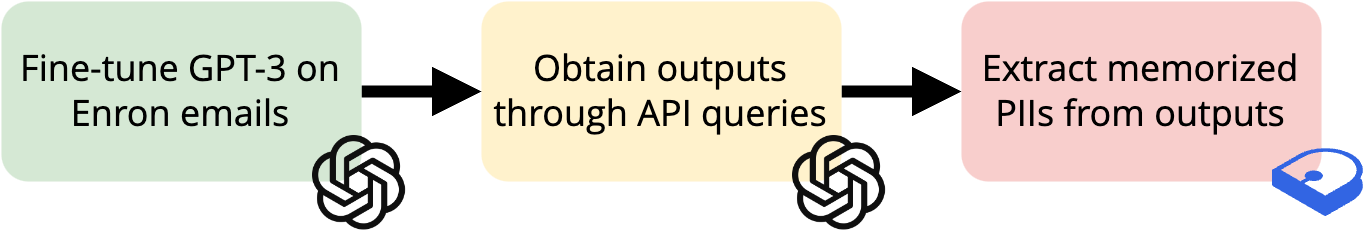}
  \caption{Step-by-step flowchart of our simulated attacks for a GPT-3 (\texttt{curie}) model trained on the Enron Email dataset using the OpenAI fine-tuning API.}
  \label{fig:pipeline}
\end{figure*}

In light of privacy concerns and regulatory interest, we investigate whether OpenAI’s fine-tuning API preserves privacy for PII. In this work, we simulate experiments where an attacker has “black box” access to a fine-tuned OpenAI language model. In this scenario, the attacker attempts to retrieve PIIs by prompting the model. A simple PII extraction attack attempts to answer the question that recent regulations pose. Specifically, we investigate whether attackers can extract PIIs that appear in the fine-tuning dataset from a language model.

We used the Enron email dataset as our training data, a gold standard in privacy research on language models for its authentic company correspondences and personal identifying information (PIIs) \cite{enron}. The open-source Enron email dataset was released to the public domain by the Federal Energy Regulatory Commission (FERC) during its 2002 investigation. It currently exists as one of the most common datasets for linguistics and privacy research because it exists as one of the only “substantial collections of real email made public” \cite{cohen}. Given that the topics in the emails discussed are business-related, attacking a language model trained on this data can reveal privacy vulnerabilities in enterprise use-cases. 

GPT-3 models excel at classification and text generation tasks \cite{gpt3}. In the following two experiments, we adapt the Enron dataset to both of these tasks, which we call \textbf{classification} and \textbf{autocomplete}. As shown in Fig \ref{fig:ov}, the first experiment uses the Enron email dataset to train an email classifier, and the second experiment uses the Enron email dataset to train an autocomplete service where a user inputs the subject of an email and the model is trained to output the body of the email. 

Our work contributes to the literature as follows: 

\begin{itemize}
    \item \textbf{Extraction attack on OpenAI fine-tuning API and GPT-3.} To our knowledge, we are the first to conduct public extraction attacks on fine-tuned GPT-3 and any proprietary fine-tuning API, including OpenAI's. OpenAI doesn't release specific information about how it fine-tunes models, so its training method (adapters, which model weights they freeze, privacy techniques) remains an black box to users. As a result, we argue that it is important for consumers to test how privacy-preserving their fine-tuning process is.
    \item \textbf{Designed practical attack scenario: autocomplete.} Besides replicating a naive extraction attack, we introduce a practical PII extraction scenario called autocomplete. This attack simulates a real-world product where a user uses the language model to suggest the body of their emails given a subject line.
\end{itemize}

\section{Relevant Works/Background}

\subsection{Recent Policy} 
% The European Union's Global Data Protection Regulation (GDPR) states that leaking personally identifiable information (PII) can result in limitation of one’s ability to exercise their rights, increased discrimination, identity theft or fraud, financial loss, and extreme economic or social disadvantage to the affected users \cite{GDPR}.
The Global Data Protection Regulation (GDPR) asserts that the leakage of personally identifiable information (PII) can lead to rights limitation, increased discrimination, identity theft, financial loss, and severe repercussion for affected users. Other significant policies regulating data leakage like China's Interim Measures for AI Services Management and the US’s Health Insurance Portability and Accountability Act are in the process of being implemented or have already been implemented.
% Other large national policies, like China's Interim Measures for the Management of Generative Artificial Intelligence Services and the United States’ Health Insurance Portability and Accountability Act, have been enacted \cite{chinese}.
% Under many of these laws, even quasi-identifiers, identifiers that can indirectly identify an individual with a combination of other information or datasets, are regulated under frameworks. 
% This includes brief identifiers like names, age, gender, ethnicity etc., revealing individuals'identity.
A detailed table with these policies can be found in Appendix section \ref{sec:appendix}. 

\subsection{Definition of Personally Identifiable Information (PII)}

We use the same definition of PII as the definition of "re-identifying information" from \cite{pilan}. Re-identifying information can fall under two categories: direct identifiers and indirect identifiers. Direct identifiers correspond to values that are unique to an identity such as a full name of person or cellphone number, whereas indirect identifiers are pieces of information that can be used in conjunction to reidentify an individual like gender and someone’s first name. If leaked, indirect identifiers can still hold large, negative privacy implications. An adversary can use the three indirect identifiers of gender, birth date, and postal code to re-identify 63-87\% of a population, given the public availability of the U.S. Census \cite{golle}.

\subsection{Fine-tuning large language models}

% Previous work such as \cite{hu2021lora} show that it is still important to fine-tune large language models from a utility/accuracy standpoint, even at the size of GPT-3. 
Earlier studies like \cite{hu2021lora} highlight the significance of fine-tuning large language models for utility and accuracy, including newer models like GPT-3. Using the OpenAI API, Fine-tuning GPT-3 starts with a large curated enterprise dataset of \texttt{prompts} and \texttt{completions}. These are two columns, where \texttt{prompts} contains all the data points and \texttt{completions} contains the ground truth classification/output/intended next step of the prompt. 
% Fine-tuning one's LLM application before deployment is highly beneficial, but can require the use of sensitive training data. For instance, to create a customer support chatbot, users may wish to fine-tune their model on confidential customer support transcripts with potentially-revealing customer details or demographics. 

\subsection{Privacy attacks}

There has been an recent increase in research on privacy attacks on machine learning models \cite{shokri2017membership, carlini2021extracting}. The former work introduces membership inference attacks (MIA), and the latter work studies data extraction attacks in the context of language models. In this study, we focus on implementing data extraction attacks on large language models, akin to \citep{he2022extracted}.
% In this paper, we focus on data extraction attacks in the context of extracted entities in large language models, similar to \citep{he2022extracted}. 
In our study, we focus on extracting personally identifiable information (PII) like \citep{lukas2023analyzing, diera2022study, mireshghallah2022memorization}. Unlike previous work that focused on language models like such as GPT-2 and BERT \cite{lukas2023analyzing, diera2022study, mireshghallah2022memorization}, we conduct an extraction attack on GPT-3 and OpenAI's fine-tuning API.

\section{Methodology}

\subsection{Classification}

Many language models are used in classification settings, such as sentiment analysis. We investigate whether an adversarial user with access to a fine-tuned GPT-3 classification model can accurately access the contents of the original fine-tuning set.

For the purpose of our initial experiment, we adopt a similar setup to recent papers that have conducted extraction attacks on smaller language models \cite{lukas2023analyzing, diera2022study}. To conduct a naive extraction attack on a GPT-3 powered classification model, we replicate the experimental setup used by \cite{diera2022study}, which conducted extraction attacks on the older transformer model BERT. In particular, we use the same subset of the original Enron email dataset used in \cite{diera2022study}. In this dataset, each email is organized within one of seven different folders (e.g., \texttt{deal discrepancies}, \texttt{personal}, \texttt{online trading}). Using the OpenAI fine-tuning API, we train models to categorize the Enron emails into their respective folders.

To enable a generative model trained for classifications to produce text, we exclude the separator (\verb|\n\n###\n\n|) used during training from the end of prompts. This separator, when placed at the end of the prompt during inference, makes the classification model return the folder name/label as the completion. However, when conducting our model extraction attacks, omitting the separator during inferences leads the GPT-3 model to generate longer completions.

We perform our attack in three steps (shown in Fig. \ref{fig:pipeline}). First, we fine-tune the GPT-3 (\texttt{curie}) model on the Enron emails dataset. Second, we 
% In the 2nd box of Fig. 2, we show an example of an email in the Enron dataset containing several PII, such as the names of individuals (“Phillip”, “Allen”, “John”, “Lavorato”) and locations (“Austin”).
query our fine-tuned model. For our extraction attack, we query the model 1800 times. For each query, we use 256 maximum token length for completion to generate around 250,000 total tokens (each token is approximately ¾ of a typical word) per model. We select a temperature value ranging from 0.5 to 1 for our generations, because we find these temperature values as a good compromise between high-quality outputs and lexical diversity. For these half of these generations, we use prompts from randomly select strings with character length 100 from an English-only subset of Common Crawl to serve as naive prompts to probe our model, inspired by the naive prompting strategy used by \cite{diera2022study}. For the other half of generations, we use a blank string prompt, similar to the strategy deployed by \cite{lukas2023analyzing}. We purposefully used random text snippets scraped from the internet as prompts to demonstrate that GPT models can be attacked without prior knowledge about the fine-tuning dataset and its contents. In order to single-out the unique, extracted words from the fine-tuned set in the next step, we also probe the GPT-3 model before fine-tuning an equal amount of times as the amount of queries we ran on the fine-tuned GPT-3 model.

Finally, we evaluate the extent to which the fine-tuned model memorized personally identifiable information (PII). In this experiment, we compute the set difference $E_{ft} - E_{base}$ to remove common PII words that would have been generated by the large language model (e.g., 'John', 'Jenny') without fine-tuning. Although this step isn't absolutely necessary, it returns us a list of words that we are more certain of being in the original fine-tuned dataset. We exclude PIIs extracted from the base model to specifically examine the memorization of PIIs during fine-tuning, disregarding any potential memorization during pre-training, in this experiment.

\subsection{Autocomplete}

In this second task, we explore email body generation, inspired by Google's Smart Compose \cite{smartcompose}. Our model suggests text for the body of an email based on the given subject.

To acquire our data, we extracted a subset of the Enron email dataset. Similar to the Annotated Enron Subject Line Corpus (AESLC) dataset from the Enron dataset \cite{zhang2019slg}, we focused on cleaning up our dataset to isolate the subjects and bodies of the emails. After randomly selecting approximately 600 emails, we manually filtered them down to a subset of 149 emails using the following filters. For the body of the emails, we applied filters such as a minimum of 3 sentences, a minimum of 25 words, and a maximum of 256 words. We manually excluded emails that seemed to be notifications, bulletins, promotions, or customer service communications, as they are less relevant to our goal of developing a "smart-compose" model for general workplace usage. Moreover, we eliminated emails that had limited natural language content, particularly those with in-line graphs or charts, as they could negatively impact the quality of the fine-tuned models' natural language outputs.

To organize our email dataset, we utilized the \texttt{prompt}/\texttt{completion} format. Given an email's subject line $s$ and body $b$, the \texttt{prompt} follows the format: "Generate the body of an email from the following subject line. Subject: [$s$]". The corresponding \texttt{completion} is: "[$b$]".

% In our analysis, we focus on “identifying” names as PIIs, either "first name + last name" or "last name + initial". When a PII is already present in the prompt and replicated in the completion, we don’t count it as a data leakage. E.g: if the subject of an email is "Meeting with Bob H.", we won't count instances of "Bob H." in the generated bodies as extracted PIIs.

The train set consists of the subjects of the emails in the training set, and the test set (henceforth labelled as OOD) are the subjects of the emails in a hold-out set of subjects that were not included in the train set. The purpose of the OOD set is to determine if these personally identifiable information (PIIs) can be unintentionally revealed when writing other emails. We aim to answer the question: If a user attempts to use this model for email composition, how likely is it for them to encounter personally identifiable information (PIIs) from other emails?

Similarly to experiment 1, we trained a GPT-3 (\texttt{curie}) model for our analysis. We conducted 5 queries per email subject using the train and test sets. Consequently, for the train set consisting of 149 email subjects, we performed a total of 745 model queries. Likewise, for the test set containing 255 email subjects, we made 1275 model queries. Qualitatively speaking, we found that a higher percentage of the generated PII were obviously Enron-specific PIIs, so we didn't apply the same set-difference PII filtering process as we did in the first experiment. 

\begin{table*}[ht]
  \centering
  \begin{tabularx}{\textwidth}{>{\centering}X|>{\centering\arraybackslash}X|>{\ttfamily\centering\arraybackslash}X}

    \toprule
    \textbf{Category} & \textbf{Number of leaked PIIs from Enron emails} & \textbf{Examples} \\
    \hline
    Person & 51 & 'Jeffrey K. Skilling', 'J. Kaminski', 'Tracy Smith', "Jeffrey C" \\
    % "Gary W", "Michael L", "Lisa B.\\
    \hline
    Organization & 92 & "Enron North America", "C-SPAN", "Enron Investor Relations", "KPMG" \\
    \hline
    Geopolitical entities (Countries, cities, states) & 28 & "Santa Clara", "Palo Alto", "Calif.", "Turkmenistan" \\
    \hline
    Facilities & 3 & "the Houston Astrodome", "Smith Street", "Enron Field" \\
    \hline
    Dollar Amounts & 28 & "19 million", "$29.95", "120,000", "$8.5 million" \\
    \hline
    Cardinal (numeric values) & 55 & "8/1/99", "5/21/2000", "11/26/99" \\
    \midrule
    Total & 257 & - \\
    \bottomrule
  \end{tabularx}
  \caption{Breakdown of leaked PIIs for classification task}
    \label{tab:leaked}

\end{table*}

\section{Results}

\subsection{Classification}

With just 1800 generations of text (around 250k tokens generated), we were able to recover 256 unique PIIs from our fine-tuning dataset. During this experiment, we extracted names of confidential corporate parties such as "Enron North America," "KPMG," and "C-SPAN," as well as real personal names like "Jeffrey K. Skilling," "J. Kaminski," "Tracy Smith," and "Jeffrey C." These extractions were made after excluding common PIIs that also appear when generating from the baseline non-fine-tuned model. Table \ref{tab:leaked} shows the breakdown of types of exposed PIIs; organization words and people names are memorized the most out of these categories. 

We observe that GPT-3 recalls 4.06\% of PII in the Enron fine-tuning dataset with a precision of 2.45\%. Precision measures an attacker's confidence that a generated PII is in the training set, and recall measures how much a PII is at risk of extraction. Therefore, approximately 1 out of every 40 PIIs retrieved from the fine-tuning dataset would be valuable for the attacker.

Through qualitative analysis, we found that a significant number of the extracted PIIs are related to the Enron Corporation and the 2000-01 California energy scandal \cite{borger}. Therefore, this extraction attack on the Enron dataset has the potential to reveal information about the fine-tuning data as well as the sensitive topics surrounding the scandal.

% Furthermore, we conducted this experiment at varying scales (100 generations, 400 generations, and 900 generations. Fig. \ref{fig:extracted_over_samples} demonstrates a positive relationship between number of generations and the number of extracted PIIs when fine-tuning GPT-3 Curie on the Enron email dataset. This suggests that as we scale our experiment larger, simulated attackers can retrieve more PIIs from the dataset. 

% \begin{figure} %[ht]
%   \centering
%   \includegraphics[width=0.5\textwidth]{figs/output.png} 
%   % [width=\textwidth]
%   \caption{The relationship between number of samples and PIIs extracted from the email dataset.}
%   \label{fig:extracted_over_samples}
% \end{figure}

\subsection{Autocomplete}

% \begin{table}
%     \centering
%     \begin{tabular}{c|c|c}
%         \hline
%         & Train & OOD \\
%         \hline
%         Number of PIIs Retrieved & 236 & 223 \\
%         \hline
%         Precision & 13.16\% & 5.71\% \\
%         \hline
%         Recall & 27.83\% & 26.29\%\\
%         \hline
%     \end{tabular}
%     \caption{Precision and recall for our data extraction attacks on the autocomplete task. We conduct a data extraction attack using email subject prompts from the training set (Train) and email subject prompts from outside the training set (OOD).
%     % Precision and recall for train and OOD datasets for autocompletion task
%     }
%     \label{tab:exp2pr}
% \end{table}

\begin{table}[ht]
\centering
\begin{tabular}{@{}cccc@{}}
\toprule
Prompts & \# PIIs retrieved & Precision & Recall \\ \midrule
Train & 236 & 13.16\% & 27.83\% \\
OOD & 223 & 5.71\% & 26.29\% \\ \bottomrule
\end{tabular}
\caption{Precision and recall for our data extraction attacks on the autocomplete task. We conduct a data extraction attack using email subject prompts from the training set (Train) and email subject prompts from outside the training set (OOD).}
\label{tab:exp2pr}
\end{table}

Our metrics in Table \ref{tab:exp2pr} reveals that a significant risk of encountering sensitive personally identifiable information (PII) exist for users of the Autocomplete machine learning model. In the case of subject lines from the training set (Train), we were able to retrieve 236 PIIs. Furthermore, for subject lines not present in the training set (OOD), we retrieved 223 PIIs. The precision for recalling PIIs was measured to be 5.71\% for train prompts and 13.16\% for OOD prompts. These results indicate that around 10\% of the PIIs emitted by the fine-tuned GPT-3 model match exact PIIs found in the Enron email training data. Additionally, fine-tuned GPT-3 recalls 26.29\% and 27.83\% of PII in the Enron fine-tuning autocomplete dataset for the train prompts and OOD prompts, respectively. With just approximately 1000 calls to fine-tuned GPT-3, we were able to identify over a quarter of the PIIs present in our dataset of about 150 emails. 

The data leakage of the OOD setup is slightly lower than the train set, but it still remains high. This means that in practical settings where a user uses the autocomplete product with novel email subjects, it remains likely that they can still see leaked PIIs when using the product.

For a comprehensive breakdown of specific examples and PII leakage, refer to Table \ref{tab:train_exp2} in Appendix section \ref{sec:appendix_breakdown} for our train set and Table \ref{tab:ood_exp2} for our OOD set.

\section{Discussion}

Our work demonstrates that sensitive personal identifying information (PII) can be extracted from both our naive setting (classification) and practical setting (autocomplete), where users have black box access to the model. We find that GPT-3 models fine-tuned for classification and autocomplete tasks can successfully retrieve sensitive PIIs with simple prompts.

Using APIs like OpenAI's fine-tuning interface exposes user-sensitive data and PII to the risk of extraction, potentially resulting in data breaches, lawsuits, and significant fines for non-compliance with privacy regulations. We believe it is valuable to explore privacy-preserving finetuning models that incorporate privacy techniques such as differential privacy and PII scrubbing to make language models resistant to extraction attacks \cite{lukas2023analyzing}. 

Recent research indicates that larger models are more prone to memorizing fine-tuning datasets and becoming vulnerable to extraction attacks \cite{lukas2023analyzing}. As OpenAI and other providers of large language models consider offering fine-tuning API support for larger models, users should be aware of the heightened susceptibility to privacy attacks and memorization if privacy mitigation techniques are not implemented.

\section*{Ethics Statement}
We have minimized public disclosure of actual names in the Enron email dataset. We have only included leaked PIIs already associated with the Enron email dataset, such as "Jeffrey Skilling", the former CEO of Enron Corporation.

\section*{Contribution Statement}
A. Sun drafted the paper, designed the experiments with E. Zemour, implemented the experiments, and made the figures. E. Zemour designed the second experiment and contributed edits to the paper. A. Saxena wrote the policy sections in the introduction, contributed research toward Table 3, and made edits to the paper. U. Vaidyanathan, E. Lin, C. Lau, V. Mugunthan provided initial direction to the paper and comments to the manuscript.

\bibliography{anthology,custom}
\bibliographystyle{acl_natbib}

\appendix

\section{Appendix}
\label{sec:appendix}

\subsection{Recent Regulations}
\label{sec:regulations}

There have been a series of recent regulations by worldwide governmental organizations toward the release of large language models and other novel generative models. These detailed examples can be found in \ref{tab:regulations}.

\begin{table*}[htbp]
    
    \centering
    % \begin{tabular}{|l|l|p{5.5cm}|}
    % \begin{tabular}{|l|l|p{\dimexpr\textwidth-2\tabcolsep-2\arrayrulewidth-0.5cm}|}
    \begin{tabular}{|p{0.15\textwidth}|p{0.23\textwidth}|p{0.57\textwidth}|}
        \hline
        \textbf{Region} & \textbf{Regulatory Example} & \textbf{Details} \\
        \hline
        USA & FTC Investigates OpenAI & The FTC has begun investigating Open AI for PII leakage \cite{ftc}. In Section 27 of the Civil Investigative Demand (CID) that the FTC sent to OpenAI, the FTC asks Open AI to explicitly to "describe in Detail all steps You have taken to address or mitigate risks that Your Large Language model Products could generate statements about individuals containing real, accurate Personal Information". \\
        \hline
        USA & Health Insurance Portability and Accountability Act (HIPAA) & HIPAA requires companies to regularly audit their system and identify potential vulnerabilities \cite{hipaa}. A failure to conduct documented risk analyses of PHI data leakage could result in breach of HIPAA. \\
        \hline
        EU & General Data Protection Regulation (GDPR) & GDPR requires companies to have a Data Protection Policy that should include technical measures to prevent data breaches (i.e. pentests): Article 5 (f) of the GDPR requires "appropriate security of the personal data, including protection against unauthorized or unlawful processing and against accidental loss, destruction or damage, using appropriate technical or organisational measures. Data leakage prevention documentation is requested when a data breach occurs: Article 88 of GDPR on the topic of data breaches, requires authorities to assess whether or not personal data had been protected by appropriate technical protection measures, effectively limiting the likelihood of identity fraud or other forms of misuse. \cite{GDPR} \\
        \hline
        China & China’s Interim Measures for the Management of Generative Artificial Intelligence Services & In recent law, the Chinese government has instituted several privacy regulations toward generative AI tools such as large language models. Article 4.4 states to "respect rights related to... privacy", and Article 7 states to "respect... privacy" "in curating training data". \cite{chinese} \\
        \hline
    \end{tabular}
    \caption{Regulatory mention of PII leakage in USA, EU, and China}
    \label{tab:regulations}
\end{table*}

\subsection{Breakdown for leaked PIIs for Autocomplete}
\label{sec:appendix_breakdown}

Specific examples and breakdown of the PII leakage for the Autocomplete task can be found in Table \ref{tab:train_exp2} for our train prompts and Table \ref{tab:ood_exp2} for our OOD prompts. 

\begin{table*}[ht]
  \centering
  \begin{tabularx}{\textwidth}{>{\centering}X|>{\centering\arraybackslash}X|>{\ttfamily\centering\arraybackslash}X}
    
    \toprule
    \textbf{Category} & \textbf{Number of leaked PIIs from train subject prompts} & \textbf{Examples} \\
    \hline
    Person & 70 & "Mark Palmer", "Janette Elbertson", "Mark Koenig" \\
    \hline
    Date & 60 & "Thursday, Sept. 14", "Wednesday, June 6, 2001" \\
    \hline
    Organization & 106 & "Charlene Jackson/Corp/Enron", "Florida Power \& Light Company", "FPL Energy" \\
    \midrule
    Total & 236 & - \\
    \bottomrule
  \end{tabularx}
  \caption{Examples of leaked people, dates, and organizations from the autocomplete task for the \textbf{train} subject prompts.}
    \label{tab:train_exp2}
  
\end{table*}

\begin{table*}[ht]
  \centering
  \begin{tabularx}{\textwidth}{>{\centering}X|>{\centering\arraybackslash}X|>{\ttfamily\centering\arraybackslash}X}
    
    \toprule
    \textbf{Category} & \textbf{Number of leaked PIIs from OOD subject prompts} & \textbf{Examples} \\
    \hline
    Person & 68 & "Elliott", "Sempra", "Lynn", "Koch" \\
    \hline
    Date & 67 & "October 3, 2001", "May 31, 2002" \\
    \hline
    Organization & 88 & "California Independent System Operator", "Enron Wholesale Services" \\
    \midrule
    Total & 223 & - \\
    \bottomrule
  \end{tabularx}
  \caption{Examples of leaked people, dates, and organizations from the autocomplete task for the \textbf{out-of-distribution (OOD)} subject prompts.}
    \label{tab:ood_exp2}
    
\end{table*}

\end{document}